\documentclass[preprint,12pt]{elsarticle}

\usepackage{lineno,hyperref}
\usepackage{amssymb}
\usepackage{graphicx}
\usepackage{amsmath}
\usepackage{subcaption}
\usepackage{color}
\usepackage{amsfonts}
\usepackage{float}
\usepackage{multirow}
 \usepackage[linesnumbered,ruled,vlined]{algorithm2e}
\usepackage{textcomp}
\usepackage{times}
\usepackage{slashbox}
\usepackage{makecell}

\journal{Image and Vision Computing}

\begin{document}

\begin{frontmatter}

\title{Patch-based Face Recognition using a Hierarchical Multi-label Matcher}


\author[]{L.~Zhang}
\ead{lzhang@34@uh.edu}
\author[]{P.~Dou}
\ead{pdou@uh.edu}
\author[]{I.A.~Kakadiaris}
\ead{ioannisk@uh.edu}
 
\address{Computational Biomedicine Lab, 4849 Calhoun Rd, Rm 373, Houston, TX 77204}

\begin{abstract}
This paper proposes a hierarchical multi-label matcher for patch-based face recognition. In signature generation, a face image is iteratively divided into multi-level patches. Two different types of patch divisions and signatures are introduced for 2D facial image and texture-lifted image, respectively. The matcher training consists of three steps. First, local classifiers are built to learn the local matching of each patch. Second, the hierarchical relationships defined between local patches are used to learn the global matching of each patch. Three ways are introduced to learn the global matching: majority voting, $\ell_1$-regularized weighting, and decision rule. Last, the global matchings of different levels are combined as the final matching. Experimental results on different face recognition tasks demonstrate the effectiveness of the proposed matcher at the cost of gallery generalization. Compared with the UR2D system, the proposed matcher improves the Rank-1 accuracy significantly by 3\% and 0.18\% on the UHDB31 dataset and IJB-A dataset, respectively.

\end{abstract}


\end{frontmatter}


\section{Introduction}
Face recognition is an active topic for researchers in the fields of biometrics, computer vision, image processing and machine learning. In the past decades, both global and local methods have been developed. Global methods learn discriminative information from the whole face image, such as subspace methods \cite{turk1991eigenfaces, belhumeur1997eigenfaces}, Sparse Representation based Classification (SRC) \cite{wright2009robust,yang2011robust} and Collaborative Representation based Classification (CRC) \cite{zhu2012multi,zhang2011sparse}. Although global methods have achieved great success in controlled environments, they are sensitive to the variations of facial expression, illumination and occlusion in uncontrolled real-world scenarios. Proven to be more robust, local methods extract features from local regions. The classic local features include Local Binary Pattern (LBP) \cite{ahonen2006face, liao2007learning}, Gabor features \cite{zhang2005local, su2009hierarchical}, Scale-Invariant Feature Transform (SIFT) \cite{luo2007person, bicego2006use} and gray value. In local methods, more and more efforts focus on patch (block) based methods, which usually involve steps of local patch partition, local feature extraction, and local matching combination. With intelligent combination, these methods weaken the influence of variant-prone or occluded patches and combine the matching of invariant or unoccluded patches. 

Based on the success of deep learning in recent years \cite{simonyan2014very, sermanet2014overfeat, zeiler2014visualizing}, many Convolutional Nerual Networks (CNNs) have been introduced in face recognition and obtained a series of breakthroughs. Effective CNNs require a larger amount of training images and larger network sizes. Yaniv \textit{et al.} \cite{taigman2014deepface} proposed to train the DeepFace system with a standard eight layer CNN using 4.4M labeled face images. Sun \textit{et al.}  \cite{sun2014deep, sun2014deep2, sun2015deepid3} developed the Deep-ID systems with more elaborate network architectures and fewer training face images, which achieved better performance. The FaceNet \cite{schroff2015facenet} was introduced with 22 layers based on the Inception network \cite{szegedy2015going, zeiler2014visualizing}. It was trained on 200M face images and achieved further improvement. Parkhi \textit{et al.} \cite{parkhi2015deep} introduced the VGG-Face network with up to 19 layers adapted from \cite{simonyan2014very}, which was trained on 2.6M images. This network also achieved comparable results and has been extended to other applications. To overcome the massive request of labeled training data, Masi \textit{et al.} \cite{masi16dowe} proposed to use domain specific data augmentation, which generates synthesis images for CASIA WebFace collection \cite{yi2014learning} based on different facial appearance variations. Their results trained with ResNet match the state-of-the-art results reported by networks trained on millions of images. Most of these methods focus on increasing the network size to improve performance. Xiang \textit{et al.} \cite{xiang2017ijcb} presented evaluation of a pose-invariant 3D-aided 2D face recognition system (UR2D), which is robust to pose variations as large as 90\textdegree{}. Different CNNs are integrated in face detection, landmark detection, 3D reconstruction and feature extraction. Eight patches are created to overcome the pose variation problem.

This paper focuses on patch-based face recognition. Although previous patch-based methods have achieved great performance, they still suffer from two drawbacks: (a) the performance is much affected by patch size and patch division, which is assigned by experimental experience and varies in different datasets. (b) since each patch is handled individually, the correlations between different patches is ignored. To overcome these drawbacks, a Hierarchical Multi-Label (HML) matcher is proposed by introducing hierarchical patch division and patch correlations. Each face image is hierarchically divided into multi-level patches for signature generation. During the matching, a local matching is obtained for each patch based on its local classifier. Then, the global matching of each patch is learned based on different types of hierarchical relationships. Last, the global matchings of different levels of patches are combined to obtain the final matching.

The contributions of this paper are improving face recognition performance by a HML-based matcher with two new techniques: (i) unifying facial patch division in face recognition, which is achieved by two ways to construct hierarchical patches. (ii) exploring the correlations between different patches based on their hierarchical relationships, which is a step usually neglected by previous methods.

Parts of this work have been published in Zhang \textit{et al.} \cite{zhang2015icb}. In this paper, it is extended by providing: (i) a hierarchical two-level patch division based on texture-lifted image; (ii) more general applications of face recognition in the wild; (iii) the evaluation on the UR2D system \cite{xiang2017ijcb}; (iv) the statistical analysis of the evaluation based on signatures from both 2D image and texture-lifted image. 

The rest of this paper is organized as follows: Section \ref{sec2} presents related work. Section \ref{sec3} describes the patch division with signatures. Section \ref{sec4} introduces the proposed matcher. The experimental design, results and analysis are presented in Section \ref{sec5}. Section \ref{sec6} concludes the paper.  
\section{Related work}
\label{sec2}


Local features computed from small patches of the face image are less likely to be corrupted than global features. Applying local features starts from the component based method, where local features are extracted and combined first. Then, classifiers are built on the combined local features. Heisele \textit{et al.} \cite{heisele2001face} introduced the component based Support Vector Machines (SVM) to avoid pose changes. Subspace models are also extended to component based methods, for example Principal Component Analysis (PCA) and Fisher Linear Discriminant (FLD) \cite{chen2004subpattern, kim2005component}. Different machine learning algorithms have been introduced into patch-based methods. Martinez \cite{martinez2002recognizing} proposed to divide face images into several local patches and model each patch with a Gaussian distribution. The final matching is reached by summing the Mahalanobis distance of each patch. Wright \textit{et al.} \cite{wright2009robust} extended SRC into a  patch version that achieves better performance by a voting ensemble. Taking into account the global holistic features, Su \textit{et al.} \cite{su2009hierarchical} developed a hierarchical method that combines both global and local classifiers. Fisher linear discriminant classifiers are applied to global Fourier transform features and local Gabor wavelet features. A two-layer ensemble is proposed to obtain the final matching. To overcome the impact of patch scale, multi-scale patch-based methods were proposed. Yuk \textit{et al.} \cite{yuk2011multi} proposed the Multi-Level Supporting scheme (MLS). First, Fisherface based classifiers are built on multi-scale patches. Then, a criteria-based class candidate selection technique is designed to fuse local matching. Zhu \textit{et al.} \cite{zhu2012multi} developed Patch-based CRC (PCRC) and Multi-scale PCRC (MPCRC). Constrained $\ell_1$-regularization is applied to combine each patch's local matching.

To learn data-driven features, Zhen \textit{et al.} \cite{lei2014learning} proposed the Discriminant Face Descriptor (DFD) that learns the most discriminant local features by minimizing the difference of the features from the same person and maximizing the difference of the features from different people. Lu \textit{et al.} \cite{lu2015cbfd} developed the Compact Binary Face Descriptor (CBFD) which learns binary codes by removing the redundancy information with unsupervised learning. Zhang \textit{et al.} \cite{ZHANG2016176} proposed a resolution-variance robust representation strategy based on LBP and Gabor features. PCA and LDA are used to reduce feature dimension. To exploit the contextural information, Duan \textit{et al.} \cite{duan2017calbfl} proposed a context-aware local binary feature descriptor by limiting the number of bitwise changes in each descriptor. The limitation of these methods is that they rely on shallow feature descriptors. 

Deep neural network based patch methods have also been developed. Mansanet \textit{et al.} \cite{MANSANET201680} proposed a model called Local Deep Neural Network (Local-DNN) for general recognition based on two key concepts: local features and deep architectures. The model learns features from small overlapping regions using discriminative feed-forward networks with several layers. Inspired by spatial pyramid pooling in image classification, Shen \textit{et al.} \cite{SHEN201694} introduced a simple and efficient feature extraction method based on pooling local patches over a multi-level pyramid. Coupled with a linear classifier, the learned features can achieve state-of-the-art performance on face recognition. 

For face recognition in an occlusion scenario, researchers also developed methods to detect and eliminate the occluded patches \cite{azeem2014survey}. Oh \textit{et al.} \cite{oh2006occlusion} introduced the Selective Local Non-Negative Matrix Factorization (S-LNMF) method. First, PCA and the Nearest Neighbor (NN) classifier are applied to detect the occluded patches. Then, LNMF-based recognition is performed on the occlusion-free patches. Zhao \textit{et al.} \cite{zhao2014occluded} proposed to partition the face image into two layers and use the difference of sparsity to detect the occluded patches. The final matching is also obtained based on the unoccluded patches. The problem with these methods is they are sensitive to the performance of occlusion detection.  

Hierarchical Multi-label Classification (HMC) is also related to the proposed framework. In HMC, each sample has more than one label and all these labels are organized hierarchically in a tree or Direct Acyclic Graph (DAG) \cite{silla2011survey, Zhang201789}. Hierarchical information in tree and DAG structures is used to improve classification performance \cite{valentini2011true, zhang2014fully}. Here, a hierarchical multi-label based matcher is introduced by making use of the hierarchical relationships between different patches to improve their local matchings. 

\section{Signatures based on hierarchical patch division}
\label{sec3}
\subsection{Signature $\mathbb{S}^{2D}$ for 2D image} 
A non-overlapping hierarchical multi-level patch division is built based on 2D face image. Let \mbox{$\mathbb{D} =\{1, 2, \cdots, D\}$} represent a set of hierarchical levels. Given a face image $X \in {{\mathbb R}}^{u\times v}$, it is set to be level $1$. First the level $1$ image is divided and obtain the level $2$ patches. Then each patch on level $2$ is divided, and obtain the patches on the next level. By dividing the patches from level $1$ to level $D-1$, all the hierarchical patches are obtained. Let $X_{i,j}\in {{\mathbb R}}^{u_{i,j}\times v_{i,j}}$ denote the $j^{th}$ patch on level $i$. Let $N$ and $N_{i}$ denote the total number of patches and the number of patches on level $i$, respectively. So $ N=\sum_{i=1}^D{N_{i}}$. Meanwhile, a hierarchical label set is defined by $\mathbb{L}={\{l}_{i,j}\}$ as the ground truth label, where $l_{i,j}$ represents the label of patch $X_{i,j}$. Note that \mbox{$l_{i,j}\in \{1,2, \cdots, C\}$}, and $C$ represents the total number of identities in the gallery. 

\begin{figure} 
	\centering
	\begin{center}
		\includegraphics[width=0.7\linewidth]{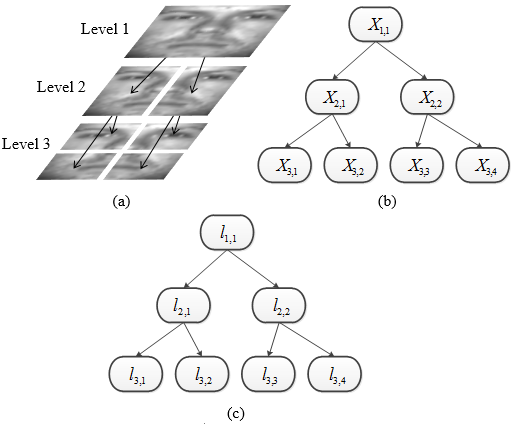}
	\end{center}
	
	\caption{An example depicted of 3-level tree-structured patch division for 2D facial image. (a) Hierarchical face division. (b) Patch hierarchy. (c) Label hierarchy.}
	\label{f1}
	
\end{figure}

By now, the 2D face image has been divided into multi-level patches and assigned a hierarchical label for each patch. Based on this patch division, any feature extraction technique can be used to generate signature $\mathbb{S}^{2D}={\{S}^{2D}_{i,j}\}$, where ${S}_{i,j}$ represent the patch-signature for the $j^{th}$ patch on level $i$. If all the patches have the same patch-signature size of $B$. The size of $\mathbb{S}^{2D}$ is $N \times B$. Figure~\ref{f1} depicts an example of a 3-level partition and its corresponding patch and label hierarchies. In practice, to emphasize local information, the level $1$ image can also start with divided patches rather than the original face image. Thus, the corresponding label hierarchy becomes a free tree without any root. 

\begin{table*}  
	\begin{center}
			\caption{ The HML patch notations with examples based on Figure~\ref{f1}.}
		\scalebox{0.7}{
			\begin{tabular}{|c|c|c|c|} \hline 
				Notations & Meanings   & Patch Relationship Examples & Label Relationship Examples \\  \hline 
				$\uparrow  (X_{i,j})$ & Parent patches of $X_{i,j}$ & $X_{1,1}= \uparrow \left(X_{2,1}\right)$ & $l_{1,1} =\uparrow \left(l_{2,1}\right)$ \\ \hline 
				$\downarrow (X_{i,j})$ & Child patches of $X_{i,j}$ & $X_{2,1}=\downarrow \left(X_{1,1}\right)$ & $l_{2,1} =\downarrow \left(l_{1,1}\right)$ \\ \hline 
				$\Uparrow  (X_{i,j} )$ & Ancestor patches of $X_{i,j}$ & $X_{1,1} =\Uparrow \left(X_{3,1}\right)$ & $l_{1,1} =\Uparrow \left(l_{3,1}\right)$ \\ \hline 
				$\Downarrow  (X_{i,j} )$ & Descendant patches of $X_{i,j}$ & $X_{3,1} =\Downarrow \left(X_{1,1}\right)$ & $l_{3,1} =\Downarrow \left(l_{1,1}\right)$ \\ \hline 
				$\Longleftrightarrow  (X_{i,j})$ & Adjacent sibling patches of $X_{i,j}$ & $X_{2,2} =\Longleftrightarrow \left(X_{2,1}\right)$ & $l_{2,2}=\Longleftrightarrow \left(l_{2,1}\right)$ \\ \hline 
				
		\end{tabular}}
		\label{ta1}
	\end{center}
\end{table*} 

By organizing multi-level patches hierarchically, the dependence between different patches can be explored based on the relationships between their labels. Following the definitions in HMC \cite{fagni2007selection, silla2011survey}, five patch notations are defined and their three hierarchical patch relationships are: ``parent-child'', ``ancestor-descendant'' and ``'adjacent siblings''. The notations and examples are shown in Table~\ref{ta1}. Sibling relationship is only defined when two patches have the same parent patch and they are adjacent. The reason of excluding non-adjacent siblings is to differentiate patches of the same level.

\subsection{Signature $\mathbb{S}^{TL}$ for texture-lifted image}

To overcome the pose variation problem, a partially overlapping tree-structured patch division is built based on texture-lifted image and integrate the partition on the patch-based UR2D system \cite{xiang2017ijcb}. Facial texture lifting is a technique that lifts the pixel values from the original 2D images to a UV map \cite{Kakadiaris2017137}. Given an original image, a 3D-2D projection matrix \cite{dou2015pose}, a 3D AFM model \cite{Kakadiaris2017}, it first generates the geometry image, each pixel of which captures the information of an existing or interpolated vertex on the 3D AFM surface. With the geometry image, a set of 2D coordinates referring to the pixels on an original 2D facial image is computed. Thus, the facial appearance is lifted and represented into a new texture image. The 3D model and a Z-Buffer technique are applied to estimate the occlusion status for each pixel. This process also generates an occlusion mask. In the UR2D system, eight patches are extracted on the texture-lifted image. Then, Deep Pose Robust Face Signature (DPRFS) is extracted for each patch \cite{xiang2017ijcb}. Due to large pose variations, some patches may be occluded. Each patch-signature contains two part: feature vector ${S}^{TL}$ with size of $512$ and a binary occlusion encoding ${O}^{TL}$, which indicates whether the patch is occluded or not. Let $\mathbb{S}^{TL}={\{S}^{TL}_{i,j}, {O}^{TL}_{i,j}\}$ represent the signature based on texture-lifted image. The $\mathbb{S}^{TL}$ signature size is $8 \times 512 + 8$.

\begin{figure} 
	\centering
	\begin{center}
		\includegraphics[width=0.8\linewidth]{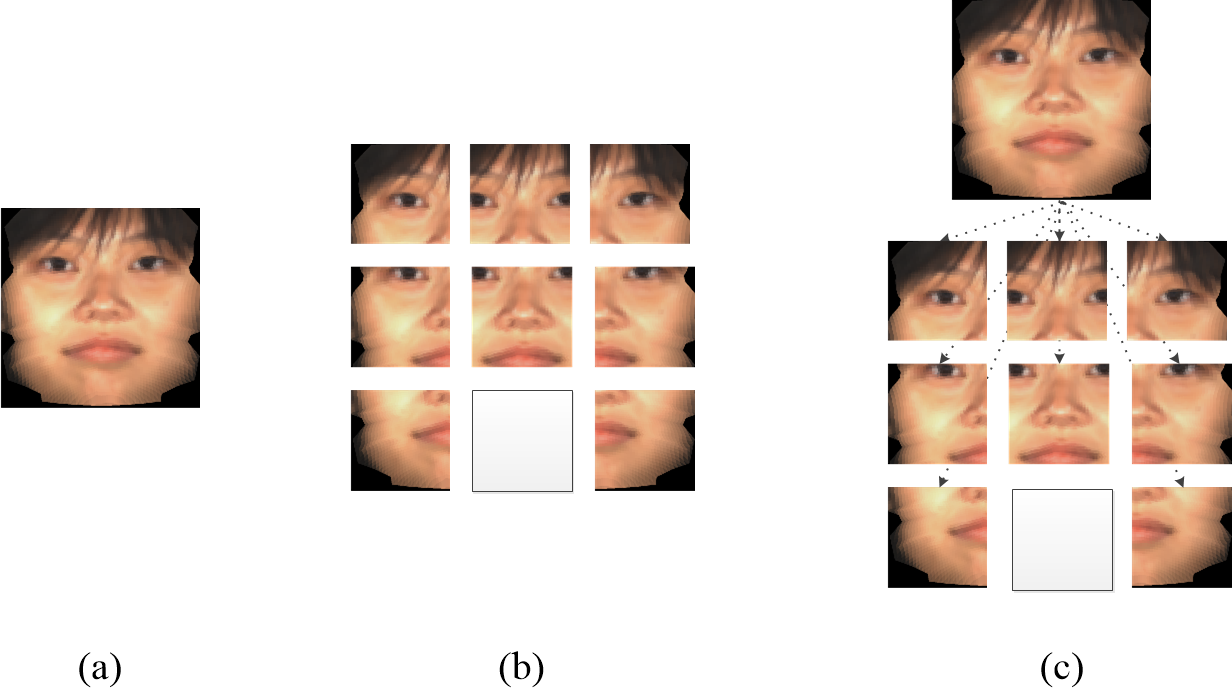}
	\end{center}
	\caption{An example depicted of 2-level tree-structured patch division for texture-lifted image. (a) Texture-lifted image. (b) Eight patch division in UR2D. (c) the patch hierarchy.}
	\label{pifr_ex}
\end{figure}

In the UR2D system, the matching score is computed by adding the cosine similarity scores of non-occluded patches directly. The major limitation is that the patch correlations are ignored. Here, based on the eight patches in the UR2D system, a two-level partially overlapping tree-structured patch division is created. An example from the UHDB31 dataset \cite{ha2017uhdb31} is shown in Figure~\ref{pifr_ex}. As with the UR2D system, the mouth patch is ignored due to expression variations. Building deeper or non-overlapping HML structures is also considered. However, DPRFS requires larger facial region to extract discriminative deep features. Following previous patch division for 2D image, the patch and label relationships are also built.

\section{Hierarchical multi-label matcher}
\label{sec4}
\subsection{Local matching}
Given a face image set \mbox{$\mathbb{X} =\left\{X^{1}, X^{2}, \cdots, X^{M} \right\}$} and its class label set  $\mathbb{Y} =\{y^{1}, y^{2},$ $ \cdots, y^{M}\}$, where \mbox{$y^{m}\in \{1,2, \cdots, C \}$} and \mbox{$m\in \{1,2, \cdots, M \}$}, the corresponding hierarchical patch sets are denoted by \mbox{$\mathbb{X}_{i,j} =\left\{X^{1}_{i,j},\ X^{2}_{i,j}, \cdots, X^{M}_{i,j}\right\}$}. As a patch-based method, local classifiers $\mathbb{F} = \{f_{i,j}(X_{i,j})\}$ are built for each patch separately. Let $\mathbb{P} = \{p^{m}_{i,j}\}$ denote the local matching set, where $p^{m}_{i,j}$ represents the matching labels of $X^{m}_{i,j}$, and \mbox{$p^{m}_{i,j}\in \{1, 2, \cdots, C\}$}. Thus:
\begin{equation} \label{GrindEQ__1_} 
p^{m}_{i,j}=f_{i,j}(X^{m}_{i,j}). 
\end{equation} 
The local matching is both signature-free and classifier-free. Any of the classifiers ({\em e.g.}, NN, SRC and CRC) can be used based on any signatures ({\em e.g.}, LBP, Gabor feature, gray value and CNN feature). For the signature of 2D image, gray value followed by CRC is evaluated. For the signature of texture-lifted image, ResNet based CNN feature followed by cosine similarity is evaluated. 

\subsection{Hierarchical global matching}
\begin{figure} 
\begin{center}
           \includegraphics[width=1\linewidth]{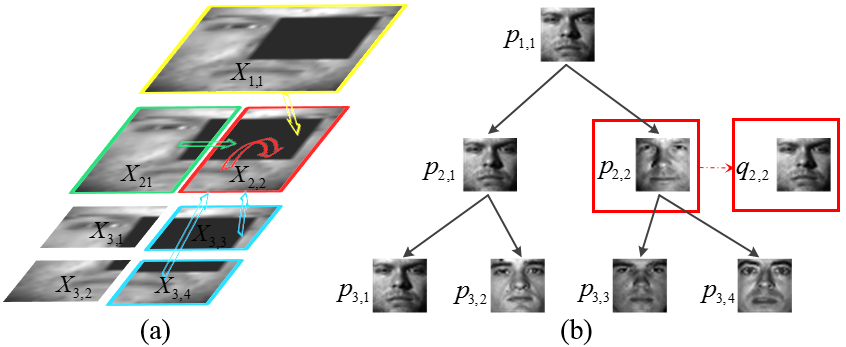}
\end{center}
   \caption{The intuition behind the proposed matcher on 2D image depicted in an occlusion scenario. (a) Tree-structured HMC patch division on an occluded image. (b) Global matching correction. The occluded region is marked by a black rectangle. For the occluded patch $X_{2,2}$ in (a), red, yellow, green, cyan arrows represent the contributions of itself, parent, adjacent sibling and child patches, respectively. In (b), it can be observed that the global matching of patch $X_{2,2}$ gives a more robust matching. }
\label{f2}
\end{figure}

\begin{figure} 
	\begin{center}
		\includegraphics[width=0.7\linewidth]{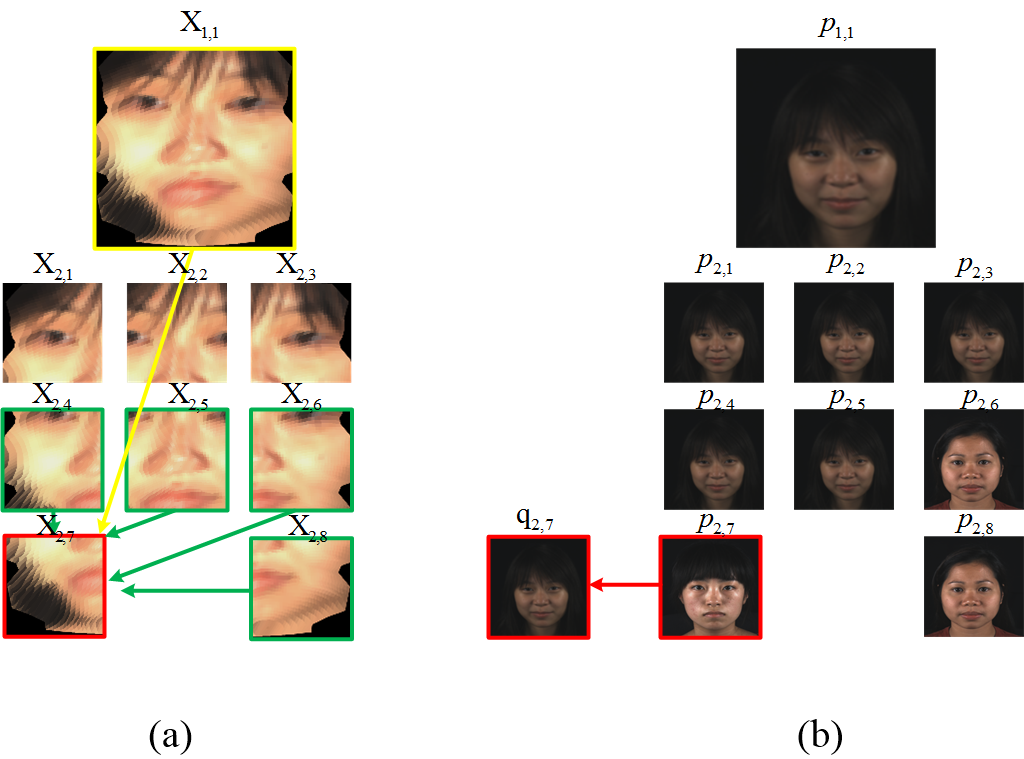}
	\end{center}
	\caption{An example of the proposed matcher on texture-lifted image depicted in large pose scenario. (a) Tree-structured HMC patch division on a texture-lifted image. (b) Global matching correction. The occluded region is due to pose variation. For the occluded patch in (a), red, yellow and green arrows represent the contributions of itself, parent and adjacent sibling patches, respectively. In (b), it can also be observed that the global matching $q_{2,7}$ corrects the local matching $p_{2,7}$.}
	\label{f22}
\end{figure}

There are several reasons why local matching is not accurate and only some patches return promising results. First, variations from facial expressions, illumination and occlusion affect different patches differently. Second, some patches are less discriminative than other patches. Third, human faces exhibit distinct structures and characteristics on different-scale patches \cite{zhu2012multi}. Previous methods usually relied on different patch sizes and ensemble methods to address these challenges. However, the correlations between different patches are neglected. In this paper, the hierarchical relationships between locally related patches are used to improve each patch's local matching and get its new label matching, which is referred as ``global matching". Let \mbox{$\mathbb{Q} =\{q^m_{i,j}\}$} denote the global matching set, where $q^m_{i,j}$ represents the global matching of  $X^m_{i,j}$, and \mbox{$q^m_{i,j}\in \{1, 2, \cdots, C\}$}.  A global classifier is defined for each patch to learn the correlation between its global matching and the local matchings of itself and its parent patches, adjacent sibling patches and child patches (if any). For patch $X_{i,j}$, its hierarchical matching matrix is defined as \mbox{ $H_{i,j} =\left[f(X_{i,j}),f({\uparrow(X_{i,j})}),f({\downarrow(X_{i,j})}),f({\Longleftrightarrow(X_{i,j})})\right] $} $\in {{\mathbb R}}^{M\times S_{i,j}}$  , where each element $h_{i,j}^{m,s}$ represents the local matching of the $s^{th}$ hierarchically related patch for the $m^{th}$ sample and $S_{i,j}$ represents the total number of hierarchically related patches. Let $\mathbb{G} = \{g_{i,j}(X_{i,j},H_{i,j})\}$ represent the learned global classifier set, so:
\begin{equation} \label{ge}
q_{i,j}^m=g_{i,j}(X_{i,j}^m,H_{i,j}^m). 
\end{equation} 
Take occlusion for example. Figure~\ref{f2} depicts the intuition of the proposed method in an occlusion scenario. It can be observed that, for the occluded patch $X_{2,2}$, three types of hierarchically related patches (parent patch $X_{1,1}$, adjacent sibling patch $X_{2,1}$ and child patches $X_{3,3}$ and $X_{3,4}$) will contribute to correct the erroneous local matching and obtain the robust global matching. Another example in a texture-lifted image is shown in Figure~\ref{f22}.

\subsubsection{Majority voting (V-HML)}

Voting is the most popular ensemble technique in patch-based methods. It is easy and training-free. The global matching of a probe patch is simply given by the majority local matching of all the hierarchically related patches. For the patches that have more than one majority matching candidate, the one that gives higher average similarity is selected. The drawback is that majority voting ignores the important differences between different hierarchically related patches. As it can be observed in the example of patch $X_{2,2}$ in Figure~\ref{f2}, compared with the child patches $X_{3,3}$ and $X_{3,4}$, the adjacent sibling patch $X_{2,1}$ and the parent patch $X_{1,1}$ provide more discriminative information. 

\subsubsection{$\ell_1$-regularized weighting (W-HML)}

Based on the above analysis, different weights are introduced to the hierarchically related patches. Let  \mbox{$\mathbf{w}_{i,j}=\{w^1_{i,j},w^2_{i,j},\cdots ,w^{S_{i,j}}_{i,j}\}^T$} represent the weight vector of the patches related to $X_{i,j}$, and \mbox{ $\sum^{S_{i,j}}_{s=1}{w^s_{i,j}}=1$}. Following \cite{zhu2012multi}, a decision matrix \mbox{$Z_{i,j} =\{z^{m,s}_{i,j}\}\in {{\mathbb R}}^{M\times S_{i,j}}$ } is defined as:
\begin{equation} \label{GrindEQ__2_} 
z^{m,s}_{i,j}=\left\{ \begin{array}{c}
+1,\ \ \ \emph{if}  \ \ y^m=h_{i,j}^{m,s}\  \\
-1,\ \ \ \emph{if}  \ \ y^m \neq h_{i,j}^{m,s}
 \end{array}.
\right.  
\end{equation} 
Note that $z^{m,s}_{i,j}=1$ means that  $h^{m,s}_{i,j}$ gives a correct matching, otherwise it gives a wrong matching. To measure the misclassification of all the hierarchically related patches, the ensemble margin of the $m^{th}$ sample can be defined as:
\begin{equation} \label{GrindEQ__3_} 
\varepsilon \left(X^{m}_{i,j} \right)=\sum^{S_{i,j}}_{s=1}{w^s_{i,j} z^{m,s}_{i,j}}.                                                                    
\end{equation} 
For the sample set $\mathbb{X}$, the ensemble loss under square loss can be defined as:
\begin{equation} \label{GrindEQ__4_} 
\begin{split} 
Loss\left({\mathbb{X}_{i,j}}\right) & =\sum^M_{m=1}{{\left[1-\varepsilon \left(X^m_{i,j}\right)\right]}^2}   \\
& =\sum^M_{m=1}{{\left(1-\sum^{S_{i,j}}_{s=1}{w^s_{i,j}z^{m,s}_{i,j}}\right)}^2}\\
& ={\left\|\mathbf{e}-Z_{i,j}\mathbf{w}_{i,j}\right\|}^2_2,                        
\end{split} 
\end{equation} 
where $\mathbf{e}=\left[ 1, 1, \cdots, 1 \right]^T$, and $dim(\mathbf{e})=S_{i,j}$. Considering that some hierarchically related patches do not make much contribution, like patch $X_{2,2}$'s adjacent sibling patch $X_{3,3}$ in Figure~\ref{f2}, the sparsity of $\mathbf{w}_{i,j}$ is enforced with $\ell_1$-norm. Also the learned weights should be positive. With these constraints, the optimization problem becomes:
\begin{equation} \label{GrindEQ__5_} 
 \begin{array}{c}
{\left\|\mathbf{e}-Z_{i,j}\mathbf{w}_{i,j}\right\|}^2_2+{\lambda \left\|\mathbf{w}_{i,j}\right\|}_1 \\ 
s.t.\ \sum^{S_{i,j}}_{s=1}{w^s_{i,j}}=1, w^s_{i,j}>0,\ s=1, 2, \cdots, S_{i,j}. \end{array}                                   
\end{equation} 
Using the same strategy as \cite{zhu2012multi}, converting the weight constraint to $\mathbf{e}\mathbf{w}_{i,j}=1$, and adding to the objective function: 
\begin{equation} \label{GrindEQ__6_} 
 \begin{array}{c}
\mathbf{w}^*_{i,j} ={argmin}_{\mathbf{w}_{i,j}}\{{\left\|\mathbf{e}'-Z'_{i,j}\mathbf{w}_{i,j}\right\|}^2_2+{\lambda \left\|\mathbf{w}_{i,j}\right\|}_1\} \\ 
s.t.\ \mathbf{w}^s_{i,j}>0,\ s=1,\ 2,\ \cdots ,S_{i,j} \end{array},                                  
\end{equation} 
where $\mathbf{e}'=\left[\mathbf{e};1\right]$, $Z'_{i,j}=\left[{Z}_{i,j};\mathbf{e}^T\right]$.  The function can be solved using popular $\ell_1$-minimization methods. After weight learning, for a testing  patch $\hat{X}^k_{i,j}$, the global matching is $\hat{q}^k_{i,j}=\arg max_c \{ \sum{w^s_{i,j}|\hat{h}^{k,s}=c \} }$.

\subsubsection{Decision rule (R-HML)}
To find the hidden relationships between different hierarchically related patches, another good method is to use rule-based classifiers \cite{safavian1991survey, breiman2001random}. The advantages include: easy to interpret and fast to generate. For the example of patch $X_{2,2}$ in Figure~\ref{f2}, a simple decision rule is:
\begin{equation} \label{GrindEQ__7_} 
\begin{split} 
& \left( f\left(  \Longleftrightarrow X_{i,j}  \right) =c\right) \  \wedge \left( f\left(\uparrow X_{i,j}\right)=c\right)\  \\
&  \longmapsto (g(X_{i,j},H_{i,j})=c ). 
\end{split}                          
\end{equation} 

\begin{figure*}
  \centering
           \includegraphics[width=1\linewidth]{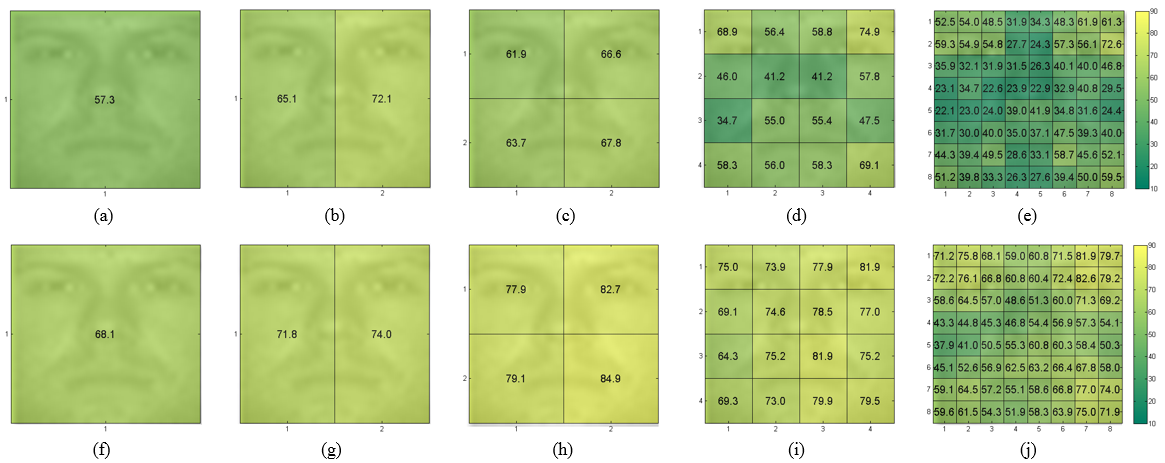}

\caption{An example depicted the per-patch accuracy comparison between local matching and global matching (\%). (a) Local level 1. (b) Local level 2. (c) Local level 3. (d) Local level 4. (e) Local level 5. (f) Global level 1. (g) Global level 2. (h) Global level 3. (i) Global level 4. (j) Global level 5.  It can be observed that, after hierarchical global learning, per patch accuracy is improved significantly on different levels.}
  \label{f3} 
\end{figure*}

Considering the variety of hierarchical relationships, Random Forest \cite{breiman2001random} is used to learn the global matching of each patch. Figure~\ref{f3} depicts an example of the comparison between local matching and global matching in the Extended Yale B dataset \cite{GeBeKr01} under the Random Forest based global classifier. A 5-level non-overlapping HML patch division is constructed on each face image. The local matching of each patch is obtained based on the simplest choices of classifier and features (NN and gray value).

\subsection{Hierarchical ensemble}
Combining multi-level global matching is an ensemble learning problem. Every method introduced above for single patch global matching learning can be applied to combine the global matchings of different levels. The proposed matcher focuses on hierarchically global learning, so the majority voting is used to get the final matching. The proposed matcher for training and deploying is summarized in Algorithms \ref{a1} and \ref{a2}, respectively. 

  \begin{algorithm}
         \caption{ The HML Matcher: Training}
        \label{a1}
    \KwIn{ Training set ${\mathbb X} =\left\{X^{1}, X^{2}, \cdots, X^{M} \right\}$ and class label set $\mathbb{Y} =\left\{y^{1}, y^{2},\cdots, y^{M}\right\}$ }
    \KwOut{ Local classifier set $\{f_{i,j}(X_{i,j})\}$, global classifier set $\{g_{i,j}(X_{i,j},H_{i,j})\}$ and final matching rule $O(\{q_{i,j}\})$}
    Partition training images hierarchically to $\{X^{m}_{i,j}\}$\\
    Build local classifier $f_{i,j}(X_{i,j})$ for each patch $X_{i,j}$\\
    \For{ $i \gets 1$ \textbf{to} $D$  } {
    \For{ $j \gets 1$ \textbf{to} $N_{i}$  }{
    Construct hierarchical matching matrix $H_{i,j}$\\
    Build global classifier $g_{i,j}(X_{i,j},H_{i,j})$\\
    Learn global matching set $\{q_{i,j}\}$ \\
    } 
    } 
    Learn final matching rule $O(\{q_{i,j}\})$\\
   \Return{\{$\{f_{i,j}(X_{i,j})\}$, $\{g_{i,j}(X_{i,j},H_{i,j})\}$, $O(\{q_{i,j}\})$\}} ;
     \end{algorithm}

   \begin{algorithm}
          \caption{ The HML Matcher: Deploying}
         \label{a2}
     \KwIn{ Probe set ${\hat{ \mathbb X}} =\left\{ \hat{X}^{1}, \hat{X}^{2}, \cdots, \hat{X}^{K} \right\}$, local classifier set $\{f_{i,j}(X_{i,j})\}$, global classifier set $\{g_{i,j}(X_{i,j},H_{i,j})\}$ and final matching rule $O(\{q_{i,j}\})$ }
     \KwOut{ Matched label set $\hat{\mathbb{Y}} =\left\{\hat{y}^{1}, \hat{y}^{2}, \cdots, \hat{y}^{K}\right\}$}
     Partition probe images hierarchically to $\{\hat{X}^k_{i,j}\}$\\
     Compute local matching $f_{i,j}(\hat{X}_{i,j})$ for each patch $\hat{X}_{i,j}$\\
     \For{ $i \gets 1$ \textbf{to} $D$  } {
     \For{ $j \gets 1$ \textbf{to} $N_{i}$  }{
     Construct hierarchical matching matrix $\hat{H}_{i,j}$\\
     Compute global matching set $\{\hat{q}_{i,j}\}$ \\
     } 
     } 
     Apply final matching rule $O(\{\hat{q}_{i,j}\})$\\    
    \Return{\{$\mathbb{\hat{Y}}$\}} ;
      \end{algorithm}

\section{Experiments}
\label{sec5}
This section evaluates the hierarchical patch division and the HML matcher with two signatures ($\mathbb{S}^{2D}$ and $\mathbb{S}^{TL}$) in different face recognition scenarios separately.   

\subsection{Signature $\mathbb{S}^{2D}$ evaluation}

The HML matcher is first evaluated on the 2D face recognition task in an occlusion scenario with the Extended Yale B dataset \cite{GeBeKr01} and the AR dataset \cite{martinez1998ar}. The Extended Yale B dataset contains 38 subjects under 9 poses and 64 illumination conditions. The image number of each subject ranges from 59 to 64. To introduce synthetic occlusion on each face image, the classic mandrill image is resized to cover 25\% of pixels at random location. The AR dataset contains over 4,000 color face images of 126 subjects with real occlusion and different facial expressions. As in \cite{zhang2011sparse,zhu2012multi}, a subset is used with both illumination and expression changes that contains 50 male subjects and 50 female subjects. Each subject has 26 images. All the face images are resized to $32 \times 32$.

\begin{figure} 
  \begin{center}
             \includegraphics[width=0.7\linewidth]{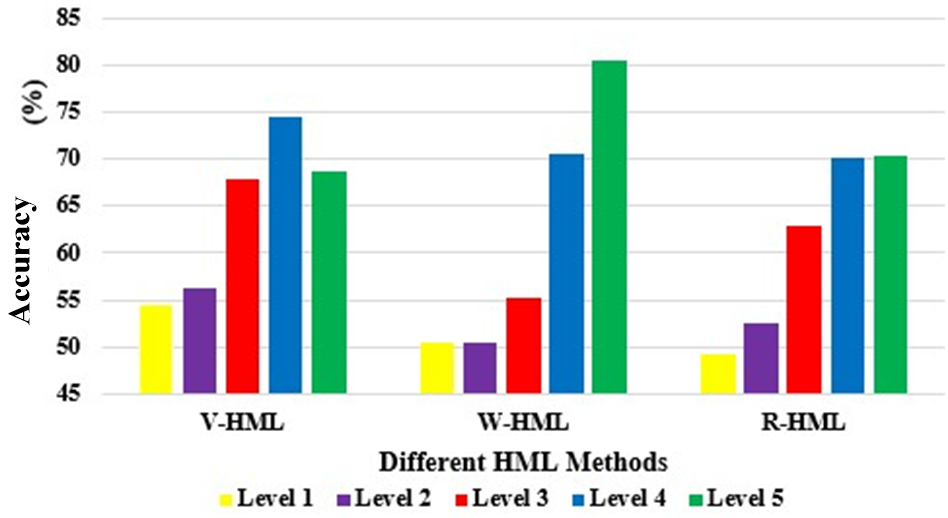}
  \end{center}
     \caption{The accuracy performance computed on the Extended Yale B dataset with different depths of HML patch division.}
  \label{nre1}
  \end{figure}

\begin{figure} 
  \begin{center}
             \includegraphics[width=0.7\linewidth]{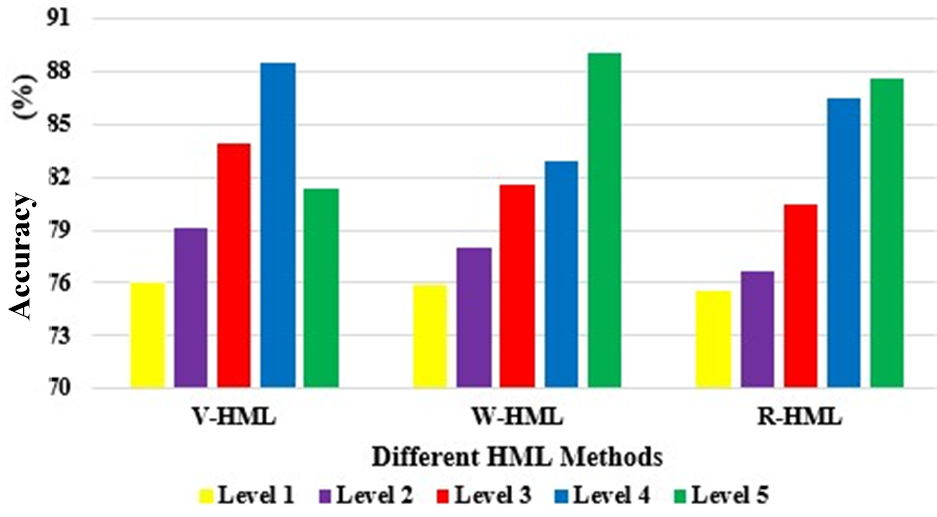}
  \end{center}
     \caption{The accuracy performance computed on the AR dataset with different depths of HML patch division.}
  \label{nre2}
  \end{figure}

\begin{figure} 
 \begin{center}
            \includegraphics[width=0.7\linewidth]{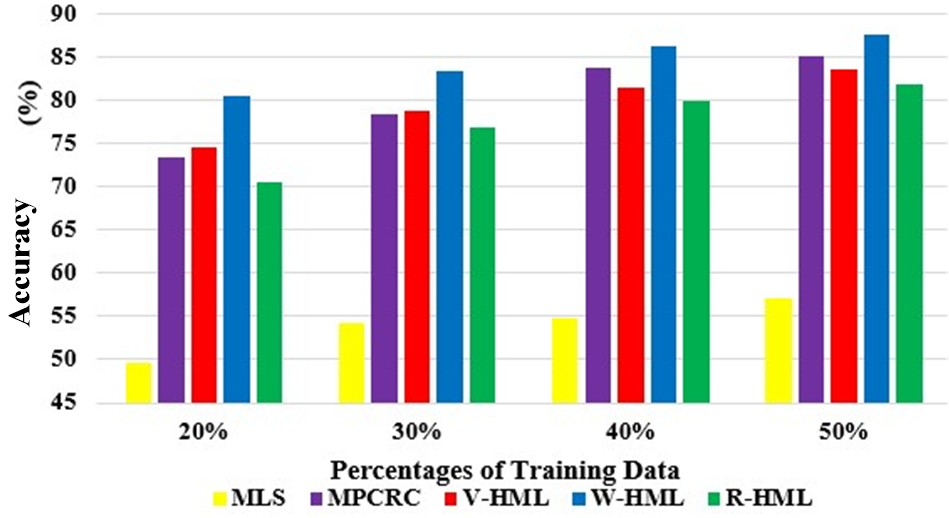}
 \end{center}
    \caption{The accuracy performance of different methods computed under different percentages of training data on the Extended Yale B dataset with $25\%$ block occlusions.}
 \label{re1}
 \end{figure}

\begin{figure} 
 \begin{center}
            \includegraphics[width=0.7\linewidth]{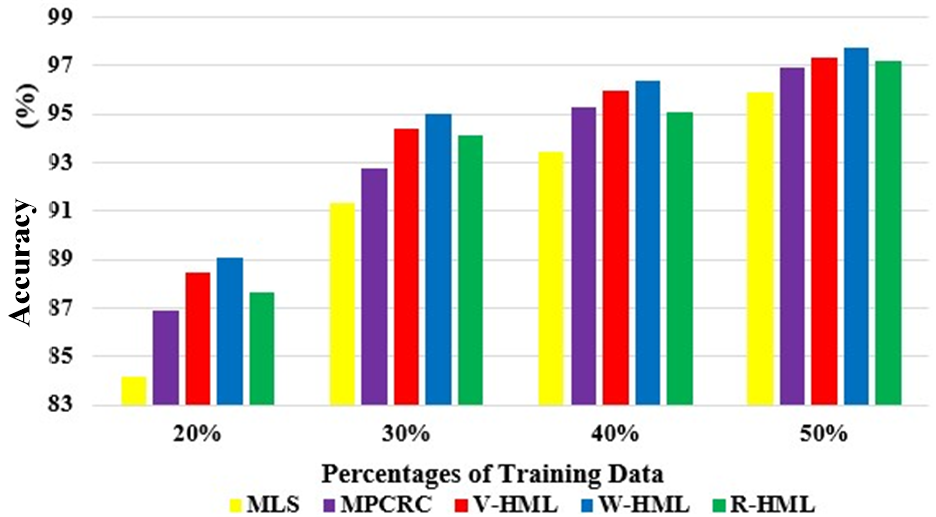}
 \end{center}
    \caption{The accuracy performance of different methods computed under different percentages of training data on the AR dataset with real occlusions.}
 \label{re2}
 \end{figure}

Based on the size of face image, a 5-level patch division is built for each image; the numbers of patches on each level are: $1 \times 1 $, $1 \times 2 $, $2 \times 2 $, $4 \times 4 $, $8 \times 8 $, respectively. According to the best result from \cite{zhu2012multi}, CRC is chosen as the local classifier. In CRC, the regularization parameter is set to 0.001. V-HML is used to represent the voting based method. In W-HML, the parameter $\lambda$ is set to 0.1. In R-HML, the number of trees is set to 150. The baseline methods are MLS and MPCRC. In MLS, the parameter $\omega$ is set to 0.1. Gray value is used as the original feature. PCA is applied to reduce dimensionality of each patch to 100 (the smaller size patches with less than 100 pixels keep their original features). To ensure fairness across subjects, The greatest common number of images are first selected randomly for each subject. Then different proportions of the selected images are used for training and testing. All the experiments were run 10 times. The influence of $D$ is first tested with 20\% of training data. The performance is depicted in Figures~\ref{nre1} and \ref{nre2}. The results of different methods are presented in Figures~\ref{re1} and \ref{re2} ($D$ is set to 4 in V-HML and 5 in W-HML and R-HML).

It can be observed in Figures~\ref{nre1} and \ref{nre2} that different HML methods achieve the best performance at different levels. V-HML performs best with 4-level path division, while W-HML and R-HML perform best with 5-level path division. In Figure~\ref{re1} it can be observed that, in the Yale B dataset, W-HML performs better than other methods under different percentages of training data. V-HML achieves better or comparable results compared to R-HML, MPCRC and MLS. The results of V-HML under 40 \% and 50 \% of training data (81.37 \%, 83.47\%) are slightly below than that of MPCRC (83.81\%, 85.03\%). In Figure~\ref{re2}, similar results can be observed in the AR dataset. W-HML achieves the best results under different percentages of training data. V-HML performs better than R-HML and other methods.  

\subsection{Signature $\mathbb{S}^{TL}$ evaluation}
\begin{figure} 
\centering
\begin{center}
	\includegraphics[width=1\linewidth]{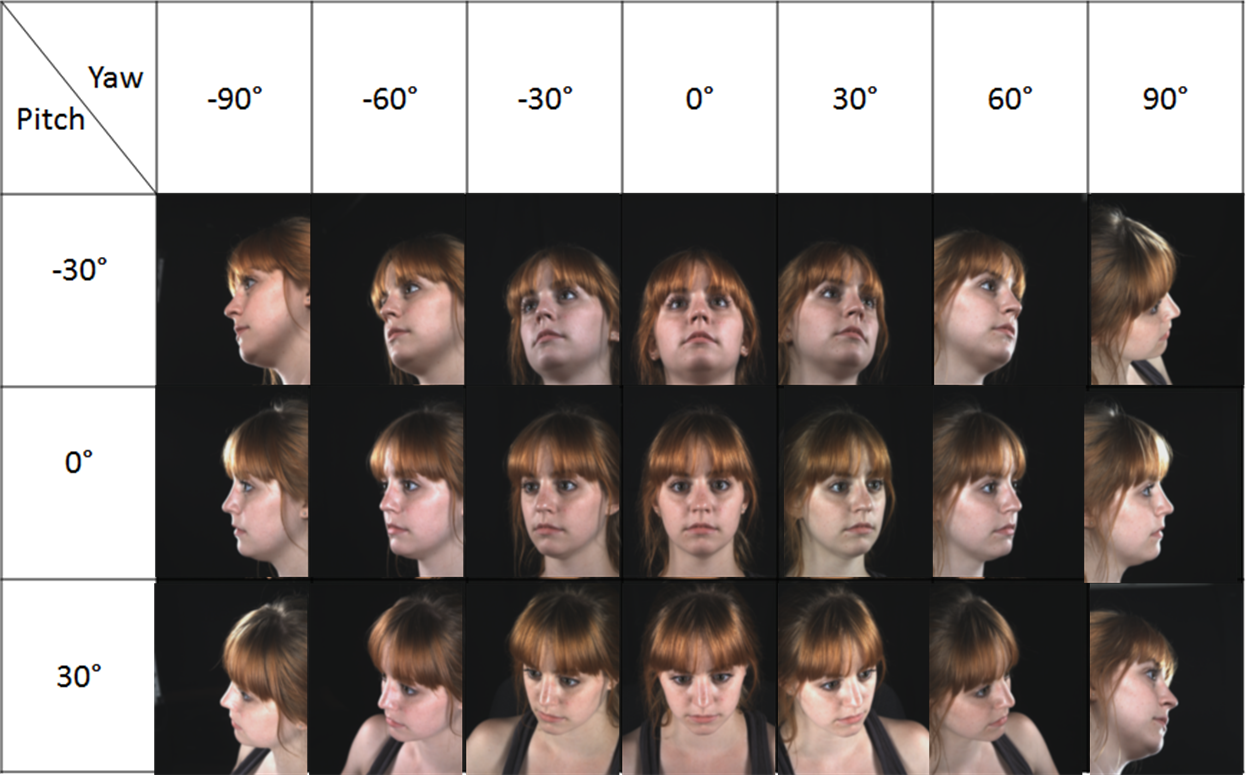}
\end{center}
\caption{Depicted image examples of different poses in the UHDB31 dataset.}
\label{uhdb31_ex}
\end{figure}

The evaluation of the proposed HML matcher on the UR2D system is evaluated with two types of face recognition scenarios: constrained environment and unconstrained environment. The datasets used for testing are the UHDB31 dataset \cite{ha2017uhdb31, zhang2017ijcb} and the IJB-A dataset \cite{klare2015pushing, zhang2018icb}. The UHDB31 dataset contains 29,106 color face images of 77 subjects with 21 poses and 18 illuminations. To exclude the illumination changes, a subset with natural illumination is selected for evaluation. To evaluate the performance of cross pose face recognition, the frontal pose (pose-11) face images are used as gallery and the remaining images from 20 poses are used as probe. Figure~\ref{uhdb31_ex} shows the example images from different poses. The IJB-A dataset \cite{klare2015pushing} contains images and videos from 500 subjects captured from ``in the wild'' environments. This dataset merges images and frames and provides evaluations on the template level. A template contains one or several images/frames of one subject. According to the IJB-A protocol, it splits galleries and probes into 10 splits. In this experiment, the same modification as \cite{xiang2017ijcb} is followed for use in close-set face recognition. The latest UR2D is used as a baseline pipeline with both PRFS and DPRFS features \cite{dou2015pose}. In addition, the results are also compared with VGG-Face, FaceNet, COTS v1.9 and ResNet \cite{xiang2017ijcb, masi16dowe, wen2016discriminative}. The performance of FaceNet on IJB-A is ignored due to identity conflicts.

To create a training set, synthetic images are generated in the UHDB31 dataset. Each gallery image is rotated, masked and cropped to create 150 images. Then, half of them are used as sub-gallery and the other half is used as sub-probe. In the IJB-A dataset, sub-gallery and sub-probe sets are also created based on gallery images. The sub-sets are used to train the HML matcher. The local matching of each patch is computed based on the DPRFS feature and cosine similarity matching. 

\begin{figure*}
	\centering
	\begin{subfigure}[b]{0.49\textwidth}
		\includegraphics[width=\textwidth]{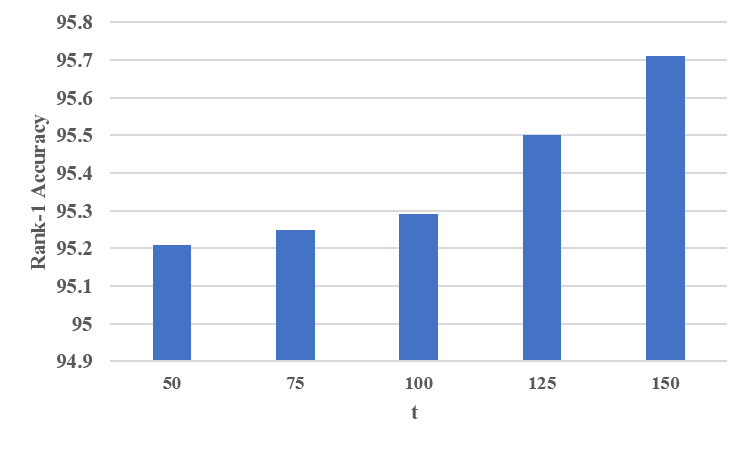}
		\caption{ }
		\label{fig:gull}
	\end{subfigure}%
	\begin{subfigure}[b]{0.49\textwidth}
		\includegraphics[width=\textwidth]{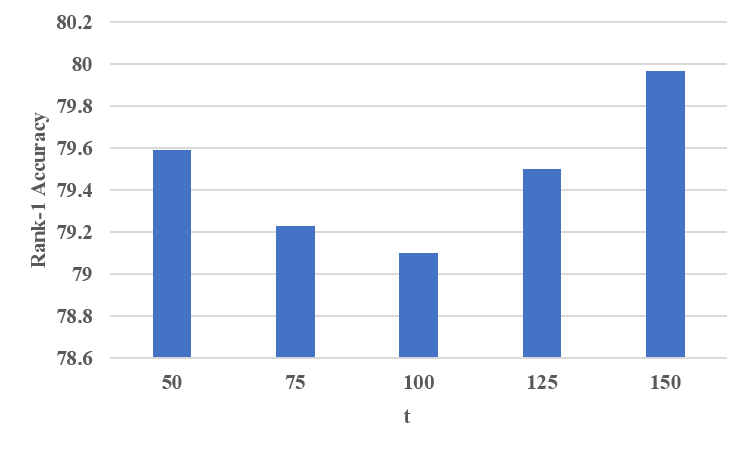}
		\caption{ }
		\label{fig:tiger}
	\end{subfigure}
	\caption{ The sensitivity of the number of trees computed in R-HML. (a) UHDB31. (b) IJB-A.}\label{re-2-p-f1}
\end{figure*}

In this more challenging task, the V-HML and W-HML failed to obtain good performance due to large poses in constrained environment and variations in unconstrained environment. Here the results are reported based of R-HML. Table~\ref{re-2-p-t1} and Table~\ref{re-2-p-t2} show the results on the two datasets. The number of tree is set to 150. The sensitivity analysis of the number of trees is shown in Figure~\ref{re-2-p-f1}.

%
%

\begin{table*}
	\begin{center}
	\caption{Rank-1 performance of different methods computed on the UHDB31 dataset (\%). The methods are ordered as VGG-Face, COTS v1.9, FaceNet, ResNet, UR2D-PRFS, UR2D-DPRFS and R-HML.}
	\vspace{-7px} 
		\scalebox{0.6}{ 
	\begin{tabular}{| c|c |c| c |c |c| c| c|} 
	\hline 
	
    \backslashbox{Pitch}{Yaw}
    & -90\textdegree{} &-60\textdegree{} &-30\textdegree{} &0\textdegree{} & +30\textdegree{} &+60\textdegree{} &+90\textdegree{} \\
    \hline
    +30\textdegree{} & 
    \makecell{14,11,58, 70,\\ 48,{\bf 82},{\bf 82}} & 
    \makecell{69,32,95, {\bf 99},\\ 90,{\bf 99},{\bf 99}} & 
    \makecell{94,90,{\bf 100}, {\bf 100}, \\ {\bf 100},{\bf 100},{\bf 100}} & 
    \makecell{99,{\bf 100},{\bf 100}, {\bf 100}, \\ {\bf 100},{\bf 100},{\bf 100}} & 
    \makecell{95,93,99, {\bf 99},\\ {\bf 100},99,99} & 
    \makecell{79,38,92, 96,\\ 95,{\bf 99},{\bf 99}} & 
    \makecell{19,7,60, 57,\\ 47,75,{\bf 77} }   \\
        \hline
    0\textdegree{} & 
    \makecell{22,9,84,94,  \\ 79,96,{\bf 98}} & 
    \makecell{88,52,99, {\bf 100},\\ {\bf 100},{\bf 100},{\bf 100}} & 
    \makecell{{\bf 100},99,{\bf 100}, {\bf 100},  \\ {\bf 100},{\bf 100},{\bf 100}} & - & 
    \makecell{{\bf 100},{\bf 100},{\bf 100}, {\bf 100}, \\ {\bf 100},{\bf 100},{\bf 100}} & 
    \makecell{94,73,99, {\bf 100},\\ {\bf 100},{\bf 100},{\bf 100}} & 
    \makecell{27,10,91,88,  \\ 84,{\bf 96},{\bf 96}}    \\
        \hline
        -30\textdegree{} & 
    \makecell{8,0,44,49, \\ 43,75,{\bf 77}} & 
    \makecell{2,19,80, {\bf 97},\\ 90,{\bf 97},{\bf 97}} & 
    \makecell{91,90,99,  {\bf 100}, \\ 99,{\bf 100},{\bf 100}} & 
    \makecell{96,99,99, {\bf 100},\\ {\bf 100},{\bf 100},{\bf 100}} & 
    \makecell{96,98,97,{\bf 100},\\ 99,{\bf 100},{\bf 100}} & 
    \makecell{52,15,90, 96,\\ 95,96,{\bf 99}} & 
    \makecell{9,3,35, 47,\\ 58,{\bf 79},{\bf 79}}    \\
    \hline
	\end{tabular}}
	\label{re-2-p-t1}
	\end{center}
\end{table*}

\begin{table*}
	\begin{center}
		\caption{The Rank-1 performance of R-HML computed on the IJB-A dataset (\%).}
		\vspace{-7px}
		\scalebox{0.65}{
			\begin{tabular}{ l |c c c c c c c c c c c} 
				\hline 
				Methods &split-1 & split-2 &split-3 &split-4 &split-5 & split-6 & split-7 &split-8 & split-9 & split-10 & Average\\
				\hline 
				    VGG-Face &76.18 &74.37 &24.33 &47.67 &52.07 &47.11 &58.31 &54.31 &47.98 &49.06 &53.16 \\
				    COTS v1.9 &75.68 &76.57 &73.66 &76.73 &76.31 &77.21 &76.27 &74.50 &72.52 &77.88 &75.73\\
				    ResNet &77.15 &74.61 &75.11 &76.91 &75.25 &77.06 &78.48 &76.84 &73.83 &77.07 &76.23\\
				    	\hline 
				    UR2D-PRFS &49.01	&49.57	&48.22	&47.75	&48.85	&44.46	&52.46	&48.22	&43.48	&48.79	&48.08\\
				UR2D-DPRFS  &78.78&77.60&{\bf	77.94 }&{\bf	79.88}&	78.44&	80.57&	81.78&	79.00&	75.94&	79.22&	78.92 \\
				R-HML  &{\bf 79.52} &{\bf 77.62}& 	77.34 & 79.58 &{\bf 78.82}&{\bf 80.78}&{\bf 81.92}&{\bf 	79.30}&	{\bf 76.37}&{\bf 	79.72}&{\bf 	79.10} \\

				\hline 
		\end{tabular}} 
		\label{re-2-p-t2}
	\end{center}
\end{table*}

From the results of Table~\ref{re-2-p-t1}, it can be observed that, the R-HML matcher can improve the performance of 4 poses and maintain the excellent performance on other poses. From Table~\ref{re-2-p-t2}, improvements can also be observed on most of the splits. From Figure~\ref{re-2-p-f1} it can be observed that different performance is obtained with different number of trees. The parameter is learned from the training sets. The limitation of the proposed models is that they require re-training based on new gallery images.

\subsection{Statistical Analysis}    
\label{SA22}    
In this section, statistical analysis is first performed for the five methods (MLS, MPCRC, V-HML, W-HML, R-HML) over eight data splits in the above experiments (four from The Extended Yale B dataset and four from the AR dataset). Following Dem\v{s}ar \textit{et al.} \cite{demvsar2006statistical}, the Friedman test \cite{friedman1937use,friedman1940comparison} and the two tailed Bonferroni-Dunn test \cite{dunn1961multiple} are used to compare multiple methods over multiple datasets. Let $r_i^j$ represent the rank of the $j^{th}$ of k algorithm on the $i^{th}$ of $N$ datasets.  The Friedman test compares the average ranks of different methods, by $R_j = \frac{1}{N} \sum_i r_i^j$. The null-hypothesis states that all the methods are equal, so their ranks $R_j$ should be equivalent. The original Friedman statistic \cite{friedman1937use,friedman1940comparison}, 
\begin{equation}\label{st122}
\mathcal{X}_F^2 = \frac{12N}{k(k+1)}[\sum_j R_j^2 - \frac{k(k+1)^2}{4}],
\end{equation}
is distributed according to $\mathcal{X}_F^2$ with $k-1$ degree of freedom. Limited by its undesirable conservative property, Iman \textit{et al.} \cite{iman1980approximations} introduced a better statistic
\begin{equation}\label{st222}
F_F = \frac{(N-1)\mathcal{X}_F^2}{N(k-1)-\mathcal{X}_F^2},
\end{equation}
which is distributed according to the F-distribution with $k-1$ and $(k-1) \times (N-1)$ degrees of freedom. First the average rank of each method is computed. The results are summarized in Table ~\ref{ta_st22}. The $F_F$ statistical value based on (\ref{st222}) is computed as $133$. With five methods and eight dataset splits, $F_F$ is distributed with $5-1$ and $(5-1) \times (8-1) = 28$ degrees of freedom. The  critical value of $F(4, 28)$ for $\alpha = 0.10$ is $2.157 < 133$, so the null-hypothesis is rejected. Then, the two tailed Bonferroni-Dunn test is applied to compare each pair of methods by the critical difference:
\begin{equation}\label{st322}
CD = q_{\alpha} \sqrt{\frac{k(k+1)}{6N}},
\end{equation}
where $q_{\alpha}$ is the critical values. If the average rank between two methods is larger than critical difference, the two methods are significantly different. According to Table 5 in \cite{demvsar2006statistical}, the critical value of five methods when $p = 0.10$ is 2.241. The critical difference is computed as $CD = 2.241 \sqrt{\frac{5 \times 6}{6 \times 8}} = 1.77$. Thus, W-HML performs significantly better than MLS, MPCRC and R-HML  (the difference between W-HML and MPCRC or R-HML, $3.5-1 = 2.5 > 1.77$). V-HML performs statistically better than MLS. The average rank differences between MPCRC and V-HML or R-HML is smaller than the critical value 1.771, so they are not significantly different. 

\begin{table*}
	\begin{center}
		\caption{The average rank of each method.}
		\vspace{-7px} 
		\begin{tabular}{ l c c c  c  c} 
			\hline 
			Measurements  &MLS  &MPCRC  &V-HML  &W-HML  &R-HML\\
			\hline
			Accuracy     &5 &3.5 &2 &1 &3.5\\  
			
			\hline 
		\end{tabular}
		\label{ta_st22}
	\end{center}
\end{table*}

The Friedman test \cite{friedman1937use,friedman1940comparison} and the two tailed Bonferroni-Dunn test \cite{dunn1961multiple} are also used to compare the performance of UR2D-DPRFS(the best baseline) and R-HML on the 30 data splits (20 from UHDB31 and 10 from IJB-A). First the average ranks of UR2D-DPRFS and R-HML are 1.67 and 1.33, respectively. The $F_F$ statistical value of Rank-1 accuracy is computed as $3.64$. With two methods and 30 data splits, $F_F$ is distributed with $2-1$ and $(2-1) \times (30-1) = 29$ degrees of freedom. The critical value of $F(1, 29)$ for $\alpha = 0.10$ is $2.88 < 3.64$, so the null-hypothesis is rejected. Then, the two tailed Bonferroni-Dunn test is applied to compare the two methods by the critical difference. The critical value of two methods when $p = 0.10$ is 1.65. the critical difference is computed as $CD = 1.65 \sqrt{\frac{2 \times 3}{6 \times 30}} = 0.30$. In conclusion, under Rank-1 accuracy, R-HML performs significantly better than UR2D-DPRFS (the difference between ranks is $1.67 - 1.33 = 0.34 > 0.30$).

\section{Conclusion}
\label{sec6}
This paper presented a HML based matcher for patch-based face recognition. The proposed matcher builds multi-level patches hierarchically and uses the hierarchical relationships to improve the local matching of each patch. The proposed matcher achieved better results compared to previous methods. Compared with the UR2D system, the proposed matcher can improve the Rank-1 accuracy significantly by 3\% and 0.18\% on the UHDB31 dataset and IJB-A dataset, respectively. The limitation of the proposed matcher is gallery generalization. Because the improvement is based on fix gallery subjects. Future work will focus on how to design a data-driven and feature-driven HML division method to create different divisions based on different datasets and features.

 \section*{Acknowledgements}
 This material is based upon work supported by the U.S. Department of Homeland Security under Grant Award Number 2015-ST-061-BSH001. This grant is awarded to the Borders, Trade, and Immigration (BTI) Institute: A DHS Center of Excellence led by the University of Houston, and includes support for the project ``Image and Video Person Identification in an Operational Environment: Phase I'' awarded to the University of Houston. The views and conclusions contained in this document are those of the authors and should not be interpreted as necessarily representing the official policies, either expressed or implied, of the U.S. Department of Homeland Security.
 
\section*{References}
\bibliographystyle{elsarticle-num}
\bibliography{egbib_all_p}

\begin{thebibliography}{10}
\expandafter\ifx\csname url\endcsname\relax
  \def\url#1{\texttt{#1}}\fi
\expandafter\ifx\csname urlprefix\endcsname\relax\def\urlprefix{URL }\fi
\expandafter\ifx\csname href\endcsname\relax
  \def\href#1#2{#2} \def\path#1{#1}\fi

\bibitem{turk1991eigenfaces}
M.~Turk, A.~Pentland, Eigenfaces for recognition, Journal of Cognitive
  Neuroscience 3~(1) (1991) 71--86.

\bibitem{belhumeur1997eigenfaces}
P.~N. Belhumeur, J.~P. Hespanha, D.~Kriegman, Eigenfaces vs. {F}isherfaces:
  Recognition using class specific linear projection, IEEE Transactions on
  Pattern Analysis and Machine Intelligence 19~(7) (1997) 711--720.

\bibitem{wright2009robust}
J.~Wright, A.~Y. Yang, A.~Ganesh, S.~S. Sastry, Y.~Ma, Robust face recognition
  via sparse representation, IEEE Transactions on Pattern Analysis and Machine
  Intelligence 31~(2) (2009) 210--227.

\bibitem{yang2011robust}
M.~Yang, D.~Zhang, J.~Yang, Robust sparse coding for face recognition, in:
  Proc. Computer Vision and Pattern Recognition, Colorado Springs, CO, 2011,
  pp. 625--632.

\bibitem{zhu2012multi}
P.~Zhu, L.~Zhang, Q.~Hu, S.~C. Shiu, Multi-scale patch based collaborative
  representation for face recognition with margin distribution optimization,
  in: Proc. European Conference on Computer Vision, Florence, Italy, 2012, pp.
  822--835.

\bibitem{zhang2011sparse}
D.~Zhang, M.~Yang, X.~Feng, Sparse representation or collaborative
  representation: Which helps face recognition?, in: Proc. International
  Conference on Computer Vision, Barcelona, Spain, 2011, pp. 471--478.

\bibitem{ahonen2006face}
T.~Ahonen, A.~Hadid, M.~Pietikainen, Face description with local binary
  patterns: Application to face recognition, IEEE Transactions on Pattern
  Analysis and Machine Intelligence 28~(12) (2006) 2037--2041.

\bibitem{liao2007learning}
S.~Liao, X.~Zhu, Z.~Lei, L.~Zhang, S.~Z. Li, Learning multi-scale block local
  binary patterns for face recognition, in: Proc. International Conference on
  Biometrics, Seoul, Korea, 2007, pp. 828--837.

\bibitem{zhang2005local}
W.~Zhang, S.~Shan, W.~Gao, X.~Chen, H.~Zhang, Local gabor binary pattern
  histogram sequence ({LGBPHS}): A novel non-statistical model for face
  representation and recognition, in: Proc. International Conference on
  Computer Vision, Beijing, China, 2005, pp. 786--791.

\bibitem{su2009hierarchical}
Y.~Su, S.~Shan, X.~Chen, W.~Gao, Hierarchical ensemble of global and local
  classifiers for face recognition, IEEE Transactions on Image Processing
  18~(8) (2009) 1885--1896.

\bibitem{luo2007person}
J.~Luo, Y.~Ma, E.~Takikawa, S.~Lao, M.~Kawade, B.~Lu, Person-specific {SIFT}
  features for face recognition, in: Proc. IEEE International Conference on
  Acoustics, Speech and Signal Processing, Vol.~2, Honolulu, HI, 2007, pp.
  593--596.

\bibitem{bicego2006use}
M.~Bicego, A.~Lagorio, E.~Grosso, M.~Tistarelli, On the use of {SIFT} features
  for face authentication, in: Proc. Computer Vision and Pattern Recognition
  Workshop, New York City, NY, 2006, pp. 1--7.

\bibitem{simonyan2014very}
K.~Simonyan, A.~Zisserman, Very deep convolutional networks for large-scale
  image recognition, in: Proc. International Conference on Learning
  Representations, San Diego, CA, 2015, pp. 1--14.

\bibitem{sermanet2014overfeat}
P.~Sermanet, D.~Eigen, X.~Zhang, M.~Mathieu, R.~Fergus, Y.~LeCun, Overfeat:
  {I}ntegrated recognition, localization and detection using convolutional
  networks, in: Proc. International Conference on Learning Prepresentations,
  Banff, Canada, 2014, pp. 1--16.

\bibitem{zeiler2014visualizing}
M.~D. Zeiler, R.~Fergus, Visualizing and understanding convolutional networks,
  in: Proc. European Conference on Computer Vision, Zurich, Switzerland, 2014,
  pp. 818--833.

\bibitem{taigman2014deepface}
Y.~Taigman, M.~Yang, M.~Ranzato, L.~Wolf, Deep{F}ace: Closing the gap to
  human-level performance in face verification, in: Proc. Computer Vision and
  Pattern Recognition, Columbus, OH, 2014, pp. 1701--1708.

\bibitem{sun2014deep}
Y.~Sun, X.~Wang, X.~Tang, Deep learning face representation from predicting
  10,000 classes, in: Proc. Computer Vision and Pattern Recognition, Columbus,
  OH, 2014, pp. 1891--1898.

\bibitem{sun2014deep2}
Y.~Sun, Y.~Chen, X.~Wang, X.~Tang, Deep learning face representation by joint
  identification-verification, in: Proc. Advances in Neural Information
  Processing Systems, Montreal, Canada, 2014, pp. 1988--1996.

\bibitem{sun2015deepid3}
Y.~Sun, D.~Liang, X.~Wang, X.~Tang, Deep{ID}3: Face recognition with very deep
  neural networks, arXiv preprint arXiv:1502.00873.

\bibitem{schroff2015facenet}
F.~Schroff, D.~Kalenichenko, J.~Philbin, Face{N}et: {A} unified embedding for
  face recognition and clustering, in: Proc. Computer Vision and Pattern
  Recognition, Boston, MA, 2015, pp. 815--823.

\bibitem{szegedy2015going}
C.~Szegedy, W.~Liu, Y.~Jia, P.~Sermanet, S.~Reed, D.~Anguelov, D.~Erhan,
  V.~Vanhoucke, A.~Rabinovich, Going deeper with convolutions, in: Proc.
  Computer Vision and Pattern Recognition, Boston, MA, 2015, pp. 1--9.

\bibitem{parkhi2015deep}
O.~M. Parkhi, A.~Vedaldi, A.~Zisserman, Deep face recognition, in: Proc.
  British Machine Vision Conference, Vol.~1, Swansea, UK, 2015, pp. 1--12.

\bibitem{masi16dowe}
I.~Masi, A.~Tran, T.~Hassner, J.~T. Leksut, G.~Medioni, Do we really need to
  collect millions of faces for effective face recognition?, in: Proc. European
  Conference on Computer Vision, Amsterdam, The Netherlands, 2016, pp.
  579--596.

\bibitem{yi2014learning}
D.~Yi, Z.~Lei, S.~Liao, S.~Z. Li, Learning face representation from scratch,
  arXiv preprint arXiv:1411.7923.

\bibitem{xiang2017ijcb}
X.~Xu, H.~Le, P.~Dou, Y.~Wu, I.~A. Kakadiaris, Evaluation of a 3{D}-aided pose
  invariant 2{D} face recognition system, in: Proc. International Joint
  Conference on Biometrics, Denver, CO, 2017, pp. 446--455.

\bibitem{zhang2015icb}
L.~Zhang, S.~Shah, I.~Kakadiaris, Hierarchical multi-label framework for robust
  face recognition, in: Proc. International Conference on Biometrics, Phuket,
  Thailand, 2015, pp. 127--134.

\bibitem{heisele2001face}
B.~Heisele, P.~Ho, T.~Poggio, Face recognition with support vector machines:
  {G}lobal versus component-based approach, in: Proc. International Conference
  on Computer Vision, Vol.~2, Vancouver, Canada, 2001, pp. 688--694.

\bibitem{chen2004subpattern}
S.~Chen, Y.~Zhu, Subpattern-based principle component analysis, Pattern
  Recognition 37~(5) (2004) 1081--1083.

\bibitem{kim2005component}
T.~K. Kim, H.~Kim, W.~Hwang, J.~Kittler, Component-based {LDA} face description
  for image retrieval and {MPEG}-7 standardisation, Image and Vision Computing
  23~(7) (2005) 631--642.

\bibitem{martinez2002recognizing}
A.~M. Mart{\'\i}nez, Recognizing imprecisely localized, partially occluded, and
  expression variant faces from a single sample per class, IEEE Transactions on
  Pattern Analysis and Machine Intelligence 24~(6) (2002) 748--763.

\bibitem{yuk2011multi}
J.~S. Yuk, K.~K. Wong, R.~H. Chung, A multi-level supporting scheme for face
  recognition under partial occlusions and disguise, in: Proc. Asian Conference
  on Computer Vision, Queenstown, New Zealand, 2010, pp. 690--701.

\bibitem{lei2014learning}
Z.~Lei, M.~Pietik{\"a}inen, S.~Z. Li, Learning discriminant face descriptor,
  IEEE Transactions on Pattern Analysis and Machine Intelligence 36~(2) (2014)
  289--302.

\bibitem{lu2015cbfd}
J.~Lu, V.~E. Liong, X.~Zhou, J.~Zhou, Learning compact binary face descriptor
  for face recognition, IEEE Transactions on Pattern Analysis and Machine
  Intelligence 37~(10) (2015) 2041--2056.

\bibitem{ZHANG2016176}
J.~Zhang, Y.~Deng, Z.~Guo, Y.~Chen, Face recognition using part-based dense
  sampling local features, Neurocomputing 184 (2016) 176--187.

\bibitem{duan2017calbfl}
Y.~Duan, J.~Lu, J.~Feng, J.~Zhou, Context-aware local binary feature learning
  for face recognition, IEEE Transactions on Pattern Analysis and Machine
  Intelligence PP~(99) (2017) 1--14.

\bibitem{MANSANET201680}
J.~Mansanet, A.~Albiol, R.~Paredes, Local deep neural networks for gender
  recognition, Pattern Recognition Letters 70 (2016) 80--86.

\bibitem{SHEN201694}
F.~Shen, C.~Shen, X.~Zhou, Y.~Yang, H.~T. Shen, Face image classification by
  pooling raw features, Pattern Recognition 54 (2016) 94--103.

\bibitem{azeem2014survey}
A.~Azeem, M.~Sharif, M.~Raza, M.~Murtaza, A survey: {F}ace recognition
  techniques under partial occlusion, International Arab Journal of Information
  Technology 11~(1) (2014) 1--10.

\bibitem{oh2006occlusion}
H.~J. Oh, K.~M. Lee, S.~U. Lee, C.~H. Yim, Occlusion invariant face recognition
  using selective {LNMF} basis images, in: Proc. Asian Conference on Computer
  Vision, Hyderabad, India, 2006, pp. 120--129.

\bibitem{zhao2014occluded}
S.~Zhao, Z.~Hu, Occluded face recognition based on double layers module
  sparsity difference, Advances in Electronics 2014 (2014) 1--6.

\bibitem{silla2011survey}
C.~N. Silla~Jr, A.~A. Freitas, A survey of hierarchical classification across
  different application domains, Data Mining and Knowledge Discovery 22~(1-2)
  (2011) 31--72.

\bibitem{Zhang201789}
L.~Zhang, S.~K. Shah, I.~A. Kakadiaris, Hierarchical multi-label classification
  using fully associative ensemble learning, Pattern Recognition 70 (2017)
  89--103.

\bibitem{valentini2011true}
G.~Valentini, True path rule hierarchical ensembles for genome-wide gene
  function prediction, IEEE/ACM Transactions on Computational Biology and
  Bioinformatics 8~(3) (2011) 832--847.

\bibitem{zhang2014fully}
L.~Zhang, S.~K. Shah, I.~A. Kakadiaris, Fully associative ensemble learning for
  hierarchical multi-label classification, in: Proc. British Machine Vision
  Conference, Nottingham, UK, 2014, pp. 1--12.

\bibitem{fagni2007selection}
T.~Fagni, F.~Sebastiani, On the selection of negative examples for hierarchical
  text categorization, in: Proc. Language and Technology Conference,
  Pozna{\'n}, Poland, 2007, pp. 24--28.

\bibitem{Kakadiaris2017137}
I.~A. Kakadiaris, G.~Toderici, G.~Evangelopoulos, G.~Passalis, D.~Chu, X.~Zhao,
  S.~K. Shah, T.~Theoharis, 3{D}-2{D} face recognition with pose and
  illumination normalization, Computer Vision and Image Understanding 154
  (2017) 137--151.

\bibitem{dou2015pose}
P.~Dou, L.~Zhang, Y.~Wu, S.~K. Shah, I.~A. Kakadiaris, Pose-robust face
  signature for multi-view face recognition, in: Proc. Biometrics Theory,
  Applications and Systems, Arlington, VA, 2015, pp. 1--8.

\bibitem{Kakadiaris2017}
I.~A. Kakadiaris, G.~Passalis, G.~Toderici, M.~N. Murtuza, Y.~Lu,
  N.~Karampatziakis, T.~Theoharis, Three-dimensional face recognition in the
  presence of facial expressions: {A}n annotated deformable model approach,
  IEEE Transactions on Pattern Analysis and Machine Intelligence 29~(4) (2007)
  640--649.

\bibitem{ha2017uhdb31}
H.~Le, I.~A. Kakadiaris, {UHDB}31: {A} dataset for better understanding face
  recognition across pose and illumination variation, in: Proc. IEEE
  International Conference on Computer Vision Workshops, Venice, Italy, 2017.

\bibitem{safavian1991survey}
S.~R. Safavian, D.~Landgrebe, A survey of decision tree classifier methodology,
  IEEE Transactions on Systems, Man, and Cybernetics 21~(3) (1991) 660--674.

\bibitem{breiman2001random}
L.~Breiman, Random forests, Machine Learning 45~(1) (2001) 5--32.

\bibitem{GeBeKr01}
A.~Georghiades, P.~Belhumeur, D.~Kriegman, From few to many: {I}llumination
  cone models for face recognition under variable lighting and pose, IEEE
  Transactions on Pattern Analysis and Machine Intelligence 23~(6) (2001)
  643--660.

\bibitem{martinez1998ar}
A.~Mart{\i}nez, R.~Benavente, The {AR} face database, Computer Vision Center,
  Universitat Aut{\'o}noma de Barcelona, CVC Technical Report 24.

\bibitem{zhang2017ijcb}
L.~Zhang, I.~A. Kakadiaris, Local classifier chains for deep face recognition,
  in: Proc. International Joint Conference on Biometrics, Denver, CO, 2017, pp.
  158--167.

\bibitem{klare2015pushing}
B.~F. Klare, B.~Klein, E.~Taborsky, A.~Blanton, J.~Cheney, K.~Allen,
  P.~Grother, A.~Mah, A.~K. Jain, Pushing the frontiers of unconstrained face
  detection and recognition: {IARPA} janus benchmark {A}, in: Proc. Computer
  Vision and Pattern Recognition, Boston, MA, 2015, pp. 1931--1939.

\bibitem{zhang2018icb}
L.~Zhang, I.~A. Kakadiaris, Fully associative patch-based 1-to-n matcher for
  face recognition, in: Proc. International Conference on Biometrics,
  Queensland, Australia, 2018, pp. 1--10.

\bibitem{wen2016discriminative}
Y.~Wen, K.~Zhang, Z.~Li, Y.~Qiao, A discriminative feature learning approach
  for deep face recognition, in: Proc. European Conference on Computer Vision,
  Amsterdam, The Netherlands, 2016, pp. 499--515.

\bibitem{demvsar2006statistical}
J.~Dem{\v{s}}ar, Statistical comparisons of classifiers over multiple data
  sets, Journal of Machine Learning Research 7 (2006) 1--30.

\bibitem{friedman1937use}
M.~Friedman, The use of ranks to avoid the assumption of normality implicit in
  the analysis of variance, Journal of the American Statistical Association
  32~(200) (1937) 675--701.

\bibitem{friedman1940comparison}
\vspace{0mm}M. Friedman, A comparison of alternative tests of significance for
  the problem of m rankings, The Annals of Mathematical Statistics 11~(1)
  (1940) 86--92.

\bibitem{dunn1961multiple}
O.~J. Dunn, Multiple comparisons among means, Journal of the American
  Statistical Association 56~(293) (1961) 52--64.

\bibitem{iman1980approximations}
R.~L. Iman, J.~M. Davenport, Approximations of the critical region of the
  fbietkan statistic, Communications in Statistics-Theory and Methods 9~(6)
  (1980) 571--595.

\end{thebibliography}
\end{document}